\documentclass[10pt,twocolumn,letterpaper]{article}

\usepackage[pagenumbers]{cvpr}

\usepackage{microtype}

\usepackage{booktabs}
\usepackage{multirow}
\usepackage{makecell}
\usepackage{pifont}
\usepackage{xspace}
\usepackage{tikz}
\usetikzlibrary{arrows.meta,positioning,calc,fit,backgrounds}

\newcommand{\ours}{FANE\xspace}
\newcommand{\doomstate}{DoomState\xspace}

\newcommand{\srule}{s^{\text{rule}}}
\newcommand{\sgeo}{s^{\text{geo}}}

\definecolor{cvprblue}{rgb}{0.21,0.49,0.74}
\usepackage[pagebackref,breaklinks,colorlinks,allcolors=cvprblue]{hyperref}

\def\confName{CVPR}
\def\confYear{2027}

\title{FANE: Function-Aligned Neural Engine for Rule-Consistent Video Game Simulation}

\author{Ziyuan Zhao \qquad Yuxuan Wu \qquad Gus Xia\\
Mohamed bin Zayed University of Artificial Intelligence\\
{\tt\small \{ziyuan.zhao, yuxuan.wu, gus.xia\}@mbzuai.ac.ae}
}

\begin{document}
\maketitle
\begin{abstract}
Current video generation models can produce high-quality, game-style videos from text prompts. Game world models build on this by injecting action control into video backbones, turning generative models into interactive game simulators. While such works move toward playable world models, maintaining consistency with underlying game rules remains a key challenge, which is crucial for human gameplay. Recent works typically address this by explicitly modeling rule-based states. However, these methods depend on deterministic logic programmed by humans or LLM agents, or on full-parameter joint fine-tuning, which limits the models' generalizability and flexibility. In this paper, we propose FANE, a hierarchical architecture that aligns a neural state engine with a pretrained video backbone so that the two levels complement each other: the engine provides the rules that the video lacks, and the video provides the geometry of the map that the engine lacks. With the backbone frozen, we train only one lightweight neural state engine and lightweight adaptation modules, achieving rule-consistent game simulation. To test our method, we release DoomState, a benchmark with exact ground truth on both videos and game states from engine simulation. Experiments show that our method maintains strong rule adherence, state accuracy and visual quality on both trained maps and unseen maps compared with other control architectures. Overall, our bi-directionally aligned neural engine architecture is rule-consistent and training-efficient, and it offers a new perspective on designing playable game world models.
\end{abstract}

\section{Introduction}
\label{sec:intro}

\begin{figure*}[t]
  \centering
  \includegraphics[width=\linewidth]{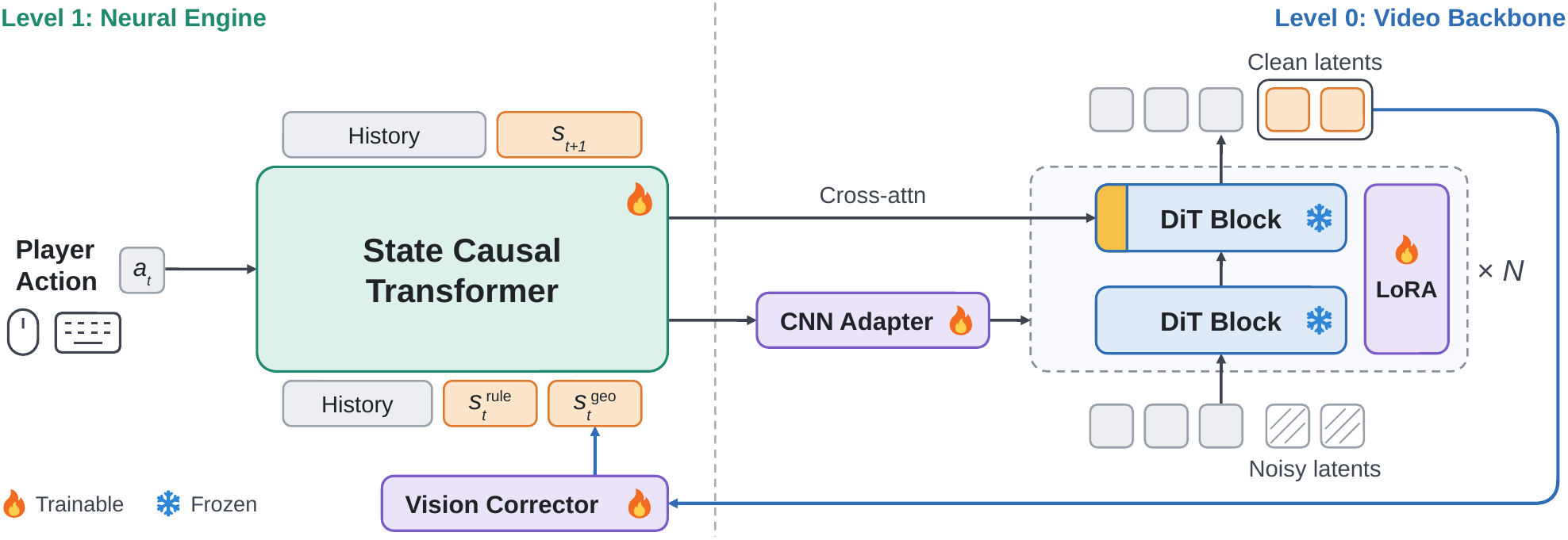}
  \caption{\textbf{Overview of \ours.}
    A causal transformer serves as the neural state engine to predict an explicit, editable game state conditioned on the player's actions. A frozen video backbone is steered through adapters, LoRA and cross attention.
    The two levels are timely and bi-directionally aligned. The state steers every video latent, and the newest rendered latents return to the engine and correct its geometry.}
  \label{fig:arch}
\end{figure*}

Video generation models can now synthesize long and realistic videos of almost any scene~\cite{wan2025,chen2025skyreels}.
The same models become interactive simulators when they are conditioned on an agent's actions: these world models predict how a scene evolves under the agent's control~\cite{ha2018worldmodels,bruce2024genie,agarwal2025cosmos,genie3,lingbot2026world}.
One application of such simulators is video game simulation, where a game world model takes the player's inputs and generates the next frames in place of the game's engine~\cite{valevski2025gamengen,alonso2024diamond,kanervisto2025wham,he2025matrixgame2,li2025gamecraft}.

To stand in for the engine, however, a game world model must keep the game's rules for as long as it runs, such that a shot spends one round, damage lowers health, and a killed monster stays dead.
An engine keeps these rules by holding the quantities they act on in a game state.
A video model keeps the state of the world only implicitly, in the frames and latents it has generated, where these quantities appear only indirectly.
Despite recent efforts toward stable long rollouts~\cite{chen2024diffusionforcing,huang2025selfforcing,xiao2025worldmem}, a stable video does not guarantee stable rules~\cite{li2026wildworld,li2026pixels}, just as generic video world models do not necessarily learn or obey physical laws~\cite{kang2025physical,motamed2026physics}. 

Recent work therefore incorporates a hierarchical design that keeps the game state explicit, as a level above the video model.
In some designs, the state is advanced by a rule-based engine or by a coding agent, and the video model serves as the renderer~\cite{zhang2026matrixgame4,huang2026pwm,wu2026boxworld}.
In others, the state is learned as a separate model that steers the video~\cite{cai2026mass,meng2026marionette}, inside a fully fine-tuned video model~\cite{lin2026stateplay,mo2026worldweaver}, or from the generated frames, which it reads back~\cite{deng2026worldmind,ye2026worldcast,lin2026gamedirector,chan2026stateworld}.
However, in 3D games, not every rule can be captured by the state alone: rules that involve the map, such as walls blocking movement and sight, need spatial information, the geometry of the map, that the state lacks, and without it the simulated game may break its rules.
Existing designs avoid this either by taking the space from a map authored in advance, or by taking it from the video at the cost of a state level with dynamics of its own.
The open question is how a learned state can keep the rules at its own level while taking the spacial information from the video.

We answer this question by keeping the learned state as its own level and linking it to the video generation backbone in both directions.
The state steers the video with the quantities it holds exactly.
The video corrects the state with the geometry of the map, such as where the player and the monsters are.
Separate levels linked in both directions are the organization proposed by function alignment~\cite{xia2025function,wu2026bridging}.
We realize this organization in the \emph{Function-Aligned Neural Engine} (\ours, \cref{fig:arch}).
The state level is a neural state engine that learns the game's state dynamics from logged play, receiving the player's actions.
The video level is a frozen pretrained video generation backbone. We allow the state engine to steer video generation through adapters and parameter-efficient fine-tuning (PEFT), and the rendered latents from video generation correct geometry information predicted by the state engine.
This design keeps training lightweight: \ours trains only a small portion of the backbone's parameters and a lightweight state engine.

To test whether a simulation keeps both the rules and the geometry, we build \doomstate on ViZDoom~\cite{kempka2016vizdoom}, with over 70 hours of gameplay, the exact state of every frame, and an independent evaluator that reads the rules from the status bar and the geometry from the monsters and the camera motion. On both trained and unseen maps, \ours keeps rule violations at the level of real video over closed-loop rollouts, following the real game's dynamics more closely than other control baselines. 
In summary, our contributions are:
\begin{itemize}
  \item A two-level formulation of game world models, in which a learned state engine keeps the rules and steers a frozen video backbone, and video generation corrects the geometric state.
  \item \doomstate, a benchmark with the exact state of every frame and an independent evaluator of both the rules and the geometry.
  \item Controlled comparisons and ablations on the same backbone, showing that \ours keeps both the rules and the geometry while training only adaptation modules to the video backbone.
\end{itemize}

\section{Related Work}
\label{sec:related}

\paragraph{Generative interactive game models.}
Neural game simulation started with models dedicated to a single game~\cite{kim2020gamegan,alonso2024diamond,valevski2025gamengen,kanervisto2025wham}.
Later work learned controls from unlabeled video~\cite{bruce2024genie} and turned pretrained video generators into playable models.
Some fine-tune the backbone~\cite{huang2026vid2world,he2025matrixgame2,li2025gamecraft}; others keep it frozen and train action modules like action cross-attention~\cite{yu2025gamefactory}, a control branch~\cite{che2025gamegenx}, modulation of the normalization layers~\cite{lingbot2026world,jiang2026olafworld}, or an adapter on the model's output~\cite{rigter2025avid}.
Diffusion forcing~\cite{chen2024diffusionforcing} and causal distillation~\cite{yin2025causvid,huang2025selfforcing} made long, streaming rollouts practical, and memory retrieval improves revisits~\cite{xiao2025worldmem}.
All of these models represent the game only in pixels, and recent analyses and datasets with ground-truth game state identify rule and state consistency as their main open problem~\cite{li2026wildworld,li2026pixels}.

\paragraph{World models with explicit state.}
Classical world models predict a compact latent state instead of pixels~\cite{ha2018worldmodels,hafner2025dreamerv3}, and early game models already split a dynamics model from a renderer~\cite{kim2020gamegan,menapace2024promptable}.
With large video generators, the state has returned in three forms.
It can be \emph{executed} where an agent-written program or hand-coded rules manage the state, and a video model renders it~\cite{zhang2026matrixgame4,huang2026pwm,wu2026boxworld,lin2026gamedirector,chan2026stateworld}.
It can be \emph{learned jointly} with the video, as a state branch~\cite{lin2026stateplay}, latent registers~\cite{mo2026worldweaver} or a latent belief~\cite{pan2025} inside a fully trained generator.
Or it can be \emph{perceived}, read back from the generated frames by perception models~\cite{deng2026worldmind,ye2026worldcast}.
Closest to \ours, two others learn the state as a separate model that only steers the renderer~\cite{cai2026mass,meng2026marionette}.
Outside games, world models predict object slots~\cite{wu2023slotformer}, 3D occupancy for driving~\cite{zheng2024occworld}, features of a pretrained encoder~\cite{zhou2025dinowm}, or depth and normals along with the video~\cite{zhen2025tesseract}.
A simulated or analytic physical state has also been coupled with a video model~\cite{liu2024physgen,wu2026bridging}.

\paragraph{Hierarchical models and alignment.}
Hierarchical world models predict or plan at several levels of abstraction~\cite{hafner2022director,lecun2022path}, usually inside one jointly trained system.
Function alignment~\cite{xia2025function} instead treats the levels as separate autoregressive processes with their own priors, aligned in time and linked in both directions.
It has been realized for symbolic music~\cite{jiang2025music} and for physical prediction~\cite{wu2026bridging}, where LoRA steers a frozen video model with analytic physical states and an additive adapter leverages the video latents to correct the analytic flow.
However, that setting is passive and its analytic flow is a fixed recursion formula. We further extend it to \emph{learned}, action-conditioned state dynamics, timely aligned with a large frozen video backbone with bi-directional connections.
On the mechanism side, aligning with a frozen generator builds on a wide range of methods like adapters for spatial conditions~\cite{zhang2023controlnet,mou2024t2iadapter}, token streams~\cite{ye2023ipadapter}, camera rays~\cite{he2025cameractrl}, and on LoRA~\cite{hu2022lora}.

\section{Method}
\label{sec:method}

\begin{figure}[t]
  \centering
  \includegraphics[width=\linewidth]{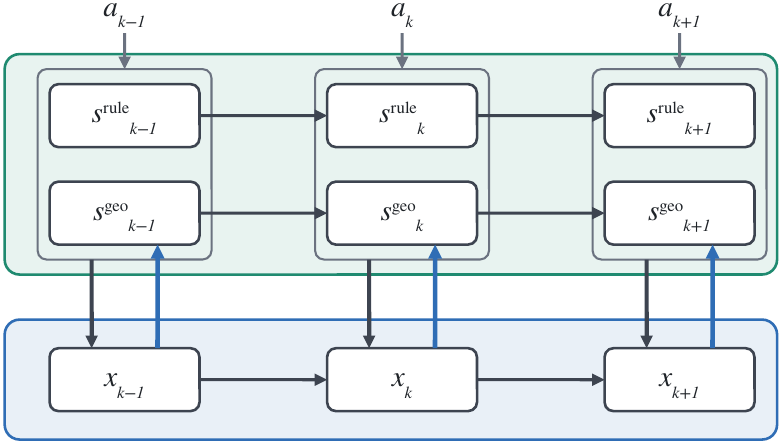}
  \caption{\textbf{Selective function alignment.}
    The engine's state (rule state $\srule$, geometry state $\sgeo$) and the backbone's latents $x$ are two timely aligned autoregressive processes.
    The full state steers every latent (top-down).
    The rendered latents correct the engine's geometry to mitigate rollout drift (bottom-up).}
  \label{fig:fa}
\end{figure}

\ours proposes a two-level game world model architecture (\cref{fig:arch}).
At the state level (Level~1), a causal transformer as \emph{neural state engine} predicts explicit game states $s_t$ for every frame $t$ conditioned on the player's actions $a_t$.
At the pixel level (Level~0), a frozen pretrained \emph{video backbone} generates the video latents $x_j$.
Each is an independent model with its own prior, and the two are coupled by a bi-directional function alignment~\cite{xia2025function} covering a top-down steering from state to video, and a bottom-up correction from video to state (\cref{fig:fa}).
In this section, we describe the engine (\cref{sec:engine}), the alignment (\cref{sec:align}) and the training (\cref{sec:training}) in detail and model implementations are given in \cref{sec:setup}.

\subsection{Neural state engine}
\label{sec:engine}

\paragraph{Explicit state.}
We categorize gameplay states into geometry states and rule states by whether they are continuous, i.e., $s_t=(\sgeo_t,\srule_t)$.
The geometry state $\sgeo_t$ holds the continuous variables such as the pose and velocity of the player and of other entities.
The rule state $\srule_t$ holds the discrete variables that are usually related to the game's logic such as health and armor.

\paragraph{State causal transformer.}
We leverage a causal transformer as neural state engine to learn the game dynamics from logged state-action sequences. The input at frame $t$ is a set of tokens including the player's state, action $a_t$, events, and entities. To aggregate different types of state data we train an extra set encoder that turns the tokens within a frame into one frame embedding $e_t$, and the causal transformer $f_\phi$ gives $h_t=f_\phi(e_{\leq t})$.
The output is a distribution $p_\phi(s_{t+1}\mid h_t)$ that factorizes over the state variables, with a separate head for each variable type, such as Gaussian for continuous variables and categorical for discrete variables. We let the action enter only the state level instead of the video level while previous works usually inject actions into both~\cite{lin2026stateplay}. Ablations about this choice will be discussed in \cref{sec:ablations}.

\subsection{Aligning the two levels}
\label{sec:align}

\paragraph{Time alignment.}
The state changes at every frame, while the backbone's autoencoder compresses $r$ frames into one latent.
We align the two on the latent's clock: $S_j=\{s_t\}_{t\in\mathcal{F}(j)}$ are the states of the frames $\mathcal{F}(j)$ of latent $j$.

\paragraph{Top-down steering.}
The video backbone of \ours is selected among pretrained video diffusion transformers (DiTs) whose weights $\theta$ stay frozen. The state enters DiT through cross-attention, CNN adapter and LoRA with parameters $\psi$. 

\emph{Time-sliced cross-attention} injects states into the backbone directly. A state encoder embeds the states of each frame as tokens and fuses the $r$ frames of latent $j$ into the token $c_j$. In every second DiT block, a new cross-attention layer lets the video tokens $z_j$ of latent $j$ attend to the two most recent state tokens:
\begin{equation}
  z_j\leftarrow z_j+\tanh(\alpha)\,\mathrm{Attn}\big(z_j,\,[c_{j-1};c_j]\big),
  \label{eq:xattn}
\end{equation}
with a zero-initialized learned gate $\alpha$ to ensure the training starts from the pretrained model.

To inform the backbone with spatial information,  the state is rasterized into screen-space condition maps at latent resolution, with channels for the occupancy and inverse depth of the entities, the camera rays of the player's motion~\cite{he2025cameractrl}, and the digit shown in each cell of the status bar.
A \emph{CNN adapter}~\cite{mou2024t2iadapter,zhang2023controlnet} adds these maps to the hidden states of several blocks.
Besides, \emph{LoRA}~\cite{hu2022lora} on the attention layers adapts the frozen model to improve visual similarity to the target game. The necessities and effects of these modules will be further discussed in the supplementary materials.

\paragraph{Bottom-up correction.}

Though the state level can predict game dynamics independently, it also can be viewed as an abstraction of the video level, for which a bottom-up correction can be beneficial by implying non-state information like environment structure.
Our \emph{vision corrector} is a small cross-attention module between the set encoder and the causal transformer, allowing the tokens of frame $t$ to attend to the newest rendered latents $X_t$.
The output is a residual correction of the frame embedding, $\tilde e_t=e_t+C_\phi(e_t,X_t)$, which is zero at initialization.
The correction acts on the geometry state only;
hence the transformer runs twice per step, with and without it, to separately output geometry and rule states:
\begin{equation}
  \sgeo_{t+1}\sim p_\phi\big(\cdot\mid f_\phi(\tilde e_{\leq t})\big),\quad
  \srule_{t+1}\sim p_\phi\big(\cdot\mid f_\phi(e_{\leq t})\big).
  \label{eq:selective}
\end{equation}

We choose such selective correction since the video level contains rich geometry information while rule states usually do not rely on game video. Restricting the correction to the geometry state reduces the risk of collapse in redundant loops.

\subsection{Training and Inference}
\label{sec:training}

Following the spirit of function alignment~\cite{xia2025function}, our two levels can be trained separately. 

\paragraph{Engine.}
The engine is trained on logged state-action sequences with the teacher-forced negative log-likelihood $\mathcal{L}_{\text{nll}}=-\sum_t\log p_\phi(s_{t+1}\mid h_t)$.
Teacher forcing suffers from exposure bias, which accumulates over a long rollout, for which we add a rollout loss.
After a logged prefix, the engine is unrolled for $H$ steps on its own reparameterized samples, and each step is scored by the likelihood of the logged displacement $\Delta\sgeo_{t+i}$:
\begin{equation}
  \mathcal{L}_{\text{roll}}=-\frac{1}{H}\sum_{i=1}^{H}w_i\,\log p_\phi\big(\Delta\sgeo_{t+i}\mid\hat h_{t+i}\big),
  \label{eq:roll}
\end{equation}
where $\hat h_{t+i}$ is computed from the unrolled states, $w_i$ is 1 while the unrolled pose stays near the logged one and 0 afterwards, and $H$ grows from 16 to 64 during training.
The engine minimizes $\mathcal{L}_{\text{eng}}=\mathcal{L}_{\text{nll}}+\lambda\,\mathcal{L}_{\text{roll}}$.
In practice we use $\lambda=0.5$. The vision corrector is trained with it on the latents of the real video. Since the latents are generated at test time, we drop the video for half of the frames and add noise to the latents to mitigate training-inference gap.

\paragraph{Video backbone.}
The adaptation modules are trained with the backbone's flow-matching objective~\cite{lipman2023flow,liu2023rectified} under per-latent noise levels $\sigma_j$, as in diffusion forcing~\cite{chen2024diffusionforcing}.
With $x^{\sigma}_j=(1-\sigma_j)\,x_j+\sigma_j\,\epsilon_j$, the loss over a set $\Omega$ of latent positions is
\begin{equation}
  \mathcal{L}_{\text{fm}}(\Omega)=\mathbb{E}_{x,\epsilon,\sigma}\big\|\big[v_{\theta,\psi}(x^{\sigma},\sigma,S)-(\epsilon-x)\big]_{\Omega}\big\|_2^2,
  \label{eq:fm}
\end{equation}
and we minimize $\mathcal{L}_{\text{vid}}=\mathcal{L}_{\text{fm}}(\Omega_{\text{all}})+\mathcal{L}_{\text{fm}}(\Omega_{\text{HUD}})$, where $\Omega_{\text{HUD}}$ are the latent rows of the heads-up display. The extra term keeps the small HUD digits legible, where autoencoder artifacts are most visible~\cite{valevski2025gamengen}.
Half of the training windows use independent noise levels, and the other half a clean, lightly noised history prefix, as in a chunked rollout.
The conditioning state is randomly sampled between the logged state and the engine's one-step prediction to improve robustness and state is also dropped at random, which allows classifier-free guidance~\cite{ho2022cfg} at inference.

\paragraph{Joint rollout.}
A rollout starts from one real frame and its state, and advances in chunks of $n$ latents. Suppose the backbone’s autoencoder compresses $r$ frames into one latent. In each chunk, the engine first reads the four most recently rendered latents to correct current geometry states, then steps the next $rn$ frames from the actions. The backbone then denoises the $n$ new latents given the clean history latents and the state tokens, and the decoder outputs the frames. Two levels achieve a joint rollout through alternating generation.

\section{Benchmark and Evaluation Protocol}
\label{sec:benchmark}

\paragraph{\doomstate.}
To build our state-grounded benchmark \doomstate, we record ViZDoom~\cite{kempka2016vizdoom} on Freedoom: Phase~2~\cite{freedoom} maps and ViZDoom scenarios at 17.5 frames per second and $384{\times}512$ pixels with the full HUD for 70 hours.
Every frame comes with the game's exact state, actions and the derived events.
The training split covers 15 maps while 5 held-out maps appear only in the test set to measure generalization. 
Episodes come from pretrained Sample Factory agents~\cite{petrenko2020samplefactory}, sticky random actions and waypoint routes that revisit places.

\paragraph{WildWorld.}
To test a second game with photorealistic graphics and a third-person camera, we use a 40-hour subset of WildWorld~\cite{li2026wildworld}, a dataset collected from the AAA action RPG Monster Hunter: Wilds.
Its state holds the tracked character and 4 entity slots, and the logged camera is used for the condition maps.

\paragraph{Tasks.}
Every rollout starts from one real frame, and all items were fixed before any model was evaluated.
\emph{Closed-loop rollout} is the main task: the model plays the recorded actions of a real episode in full closed loop of 30s length.
\emph{State editing} applies one edit to the state at 2\,s including giving ammunition, health or armor, taking ammunition, summoning a monster, or warping the player to a place visited earlier.
Its ground truth comes from re-simulating the episode in the game with and without the edit, which yields exact counterfactual video.
Both tasks are evaluated on trained maps and unseen OOD maps.
On WildWorld, the video is rendered from the logged state. 
The supplementary material adds rendering from the true state and 60-second revisit routes to further test backbone controllability and video consistency.

\paragraph{Independent evaluator.}
To test state metrics from the rendered videos, we train several independent evaluators on real frames of training episodes.
They are a template-matching HUD reader, a Faster R-CNN detector~\cite{ren2015fasterrcnn,howard2019mobilenetv3} for monsters , an ego-motion regressor on frame pairs and an event classifier.
Applied to the real video, the evaluator gives the error floor of each metric.

\paragraph{Metrics.}
\emph{Status} is the accuracy of HUD fields covering health, armor and ammunition read from the video against the real game at the same frame.
\emph{Enemy} is the per-frame error in the number of visible monsters, and \emph{Ego} the per-frame error, in map units, of the camera motion estimated from the video against the real game's.
\emph{Violations} count rule breaks per minute: ammunition dropping without a shot, health dropping without damage or rising without a pickup, and dead enemies or collected items reappearing.
For edits, \emph{compliance} is the percentage of edits for which, within 1\,s, the video agrees better with the edited re-simulation than with the unedited one.
Image quality is measured by LPIPS~\cite{zhang2018lpips} and FVD~\cite{unterthiner2018fvd}. WildWorld has no state evaluator, so it is scored by PSNR, SSIM, LPIPS and FVD only.

\section{Experiments}
\label{sec:experiments}

\begin{table*}[t]
  \centering
  \footnotesize
  \setlength{\tabcolsep}{3.4pt}
  \caption{\textbf{Closed-loop rollouts on \doomstate}.
    Train.: number of trainable parameters, including the state engine.
    Status: agreement of the HUD with the real game (\%); Enemy: error in the number of visible monsters; Ego: ego-motion error in map units per frame; Viol.: rule violations per minute.}
  \label{tab:main}
  \begin{tabular}{@{}l c cccccc c cccccc@{}}
    \toprule
    & & \multicolumn{6}{c}{Training maps} && \multicolumn{6}{c}{Unseen OOD maps} \\
    \cmidrule{3-8}\cmidrule{10-15}
    Method & Train. & Status$\uparrow$ & Enemy$\downarrow$ & Ego$\downarrow$ & Viol.$\downarrow$ & LPIPS$\downarrow$ & FVD$\downarrow$ &&
                      Status$\uparrow$ & Enemy$\downarrow$ & Ego$\downarrow$ & Viol.$\downarrow$ & LPIPS$\downarrow$ & FVD$\downarrow$ \\
    \midrule
    \multicolumn{15}{@{}l}{\emph{Frozen backbone, state-free}} \\
    Action-LoRA~\cite{yu2025gamefactory}  & 55M & 52.9 & 0.535 & 8.77 & 25.17 & 0.637 & 484 && 44.4 & 0.690 & 8.26 & 31.33 & 0.666 & 563 \\
    AVID~\cite{rigter2025avid}            & 161M & 63.3 & 0.490 & 10.86 & 35.50 & 0.560 & 730 && 58.7 & 0.678 & 9.62 & 42.92 & \textbf{0.578} & 763 \\
    \multicolumn{15}{@{}l}{\emph{Fully fine-tuned backbone}} \\
    Action-FT~\cite{he2025matrixgame2}    & 1.47B & 57.2 & 0.543 & 8.48 & 17.42 & 0.614 & 436 && 51.6 & 0.800 & 8.33 & 23.67 & 0.634 & 523 \\
    StatePlay~\cite{lin2026stateplay}     & 1.82B & 62.4 & 0.599 & 10.48 & 49.17 & \textbf{0.540} & \textbf{262} && 57.2 & 0.811 & 10.34 & 61.00 & 0.585 & \textbf{399} \\
    \multicolumn{15}{@{}l}{\emph{Frozen backbone, state-available}} \\
    State-as-Text~\cite{deng2026worldmind} & 64M & 61.8 & 0.513 & 11.01 & 27.75 & 0.634 & 683 && 54.6 & 0.578 & 10.93 & 28.83 & 0.663 & 783 \\
    Rule-Engine~\cite{zhang2026matrixgame4} & 88M & 70.5 & 2.598 & 13.59 & \textbf{3.92} & 0.555 & 509 && 67.5 & 2.453 & 11.53 & 4.33 & 0.604 & 697 \\
    \textbf{\ours (ours)}                 & 128M & \textbf{79.3} & \textbf{0.361} & \textbf{7.63} & 5.92 & 0.549 & 407 && \textbf{74.3} & \textbf{0.557} & \textbf{6.85} & \textbf{1.92} & 0.589 & 511 \\
    \bottomrule
  \end{tabular}
\end{table*}

\subsection{Setup}
\label{sec:setup}

\paragraph{Implementation.}
Our main backbone is SkyReels-V2-DF-1.3B~\cite{chen2025skyreels}, a video diffusion transformer trained with diffusion forcing~\cite{chen2024diffusionforcing}, with the Wan~2.1 autoencoder~\cite{wan2025} compressing four frames in one latent.
It stays frozen with state cross-attention added in 15 of its 30 blocks, a CNN adapter feeding 5 blocks  and LoRA of rank 64. In total, these modules introduce 6\% additional trainable parameters relative to the backbone.
The neural state engine includes a 2-layer set encoder, an 8-layer causal transformer with a 64-frame context and a  vision corrector reading the 4 newest latents.
We train the interfaces for 30k steps and the engine for 24k steps, and fine-tune the autoencoder's decoder on generated latents with reconstruction, perceptual and adversarial losses ~\cite{zhang2018lpips,isola2017pix2pix,esser2021taming}. To keep the comparison fair, all compared methods share this decoder.
For our DiT backbone, a rollout advances in chunks of 3 latents (12 frames) given 10 history latents, with 20 sampling steps and a classifier-free guidance of 1.5.

\paragraph{Baselines.}
We compare all the control architectures on the same backbone, data, sampler, 30k-step budget and decoder, and group them by whether the backbone is frozen and whether a state is available (\cref{tab:main}).
\begin{itemize}[leftmargin=*, topsep=3pt, itemsep=2pt, parsep=0pt]
  \item \emph{Frozen backbone, state-free.}
    \textbf{Action-LoRA} adds action control through new cross-attention to per-frame action tokens and LoRA, which is the action-module recipe of~\cite{yu2025gamefactory,lingbot2026world,jiang2026olafworld}.
    \textbf{AVID}~\cite{rigter2025avid} leaves the backbone untouched and trains an action-conditioned adapter on its output.
  \item \emph{Frozen backbone, state-available.}
    \textbf{State-as-Text} serializes the state into the prompt and trains LoRA, the text interface of~\cite{deng2026worldmind,lin2026gamedirector,chan2026stateworld}; in closed loop it is driven by our engine.
    \textbf{Rule-Engine} replaces our learned engine with hand-written Doom rules covering physics, weapons, damage, monster behavior, and pickups, and renders them with our adapters and LoRA.
    It stands for systems that execute the rules in code~\cite{zhang2026matrixgame4,huang2026pwm,wu2026boxworld,chan2026stateworld, lin2026gamedirector}. Like our engine, it has no access to the map file, so it can still make violations even though its dynamics are perfect.
  \item \emph{Fully fine-tuned backbone.}
    \textbf{Action-FT} fine-tunes the whole backbone on the action condition, as in~\cite{he2025matrixgame2,li2025gamecraft,huang2026vid2world}. This baseline remains state-free.
    \textbf{StatePlay}~\cite{lin2026stateplay} is state-available. It generates the video and a compact state jointly, with full fine-tuning.
    We re-implement it on our backbones and train it with the same training budget.
\end{itemize}

\subsection{Main results}
\label{sec:main}

\Cref{tab:main} reports the closed-loop task on training maps and unseen OOD maps.

\paragraph{Video quality and rule consistency.}
Over all results on closed-loop rollouts, \ours performs the best on all state-related metrics while video quality remains comparable with fully fine-tuned baselines. State-free models violate the game's rules 23.7--42.9 times per minute on unseen OOD maps, whether their backbone is frozen or fully fine-tuned, while \ours stays at a low level of violations close to the ground truth video.
Fully fine-tuned baselines behave well on LPIPS and FVD but worse on state and rule-consistency.

\paragraph{How the state is modeled matters.}
Comparisons among state-available methods show that how the state is modeled matters. State-as-Text receives the same engine state as \ours, but still breaks the rules 28.8 times per minute, implying that efficient control of the backbone with only prompt is difficult.
Rule-Engine keeps a low level of rule violations. However, without access to the real map, its enemy and ego-motion prediction drift a lot, since the information of environment is absent and only from video backbone's imagination, limiting its generalizability. Under the same training budget, StatePlay fails to exceed \ours on state metrics. This may be because of entanglement of state and video or insufficient fine-tuning. These results together indicate that our modeling of states is beneficial for state-video alignment, especially when data is limited.

\paragraph{Frozen or fine-tuned backbone.}

Action-FT trains all parameters and still breaks the rules 23.7 times per minute.
StatePlay learns its state jointly with the video and its clips have the lowest FVD, but its state metrics are worse than Action-FT and \ours. These results show that full parameter fine-tuning mainly improves video quality instead of rule consistency. Frozen methods can achieve good state-video alignment while saving training effort, but improve video quality less. Trade-off between these two paths remains to be discussed in the future.

\paragraph{Generalization.}
From training maps to unseen maps, \ours suffers from some performance drop on most metrics but still drops less and behaves better on the state compared with each baseline's results, showing that our neural engine control can capture rules behind state data and efficiently generalize to OOD tasks.

\begin{figure*}[t]
  \centering
  \includegraphics[width=\linewidth]{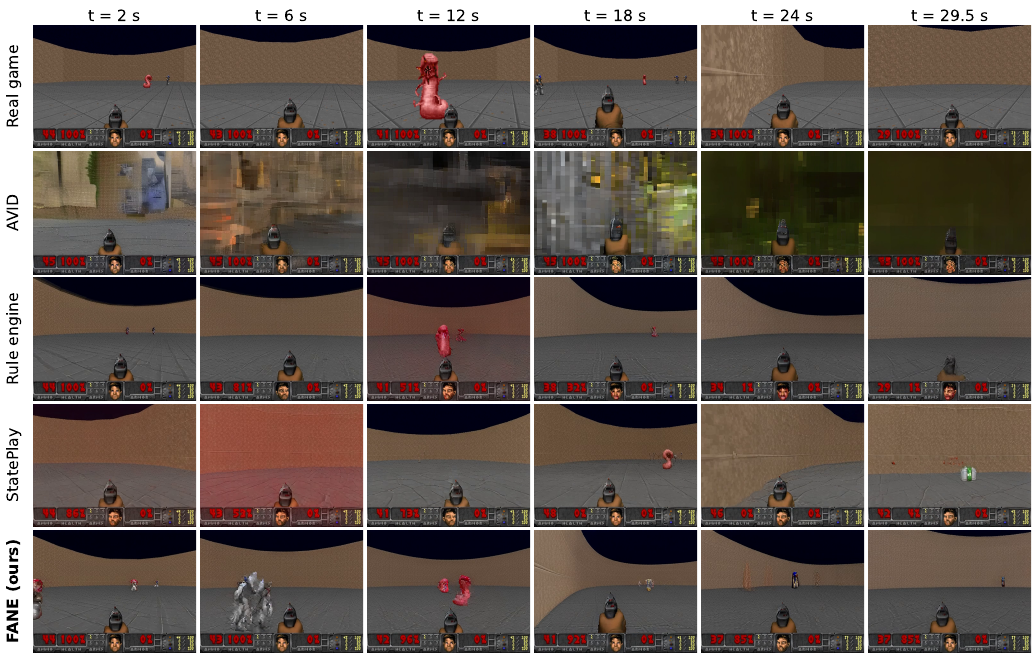}
  \caption{\textbf{Closed-loop rollouts on \doomstate} from the same first frame and the same recorded actions. The HUD shows gameplay states like ammunition and health. Compared with the baselines, \ours maintains good video and state consistency in 30s clips.}
  \label{fig:doom}
\end{figure*}

\subsection{Editability}
\label{sec:edit}

\begin{table}[t]
  \centering
  \footnotesize
  \setlength{\tabcolsep}{6pt}
  \caption{\textbf{State edits on unseen maps.}
    One edit is applied to the state at 2\,s and the ground truth is the game re-simulated with the edit.
    Compliance (\%) within 1\,s per edit type and overall.
    State-free models cannot receive an edit and can be viewed as the floor.}
  \label{tab:edits}
  \begin{tabular}{@{}lccccc@{}}
    \toprule
    Method & Give & Take & Summon & Warp & All$\uparrow$ \\
    \midrule
    Action-LoRA & 9.1 & 25.0 & \textbf{25.0} & 50.0 & 21.7 \\
    AVID & 9.1 & 0.0 & \textbf{25.0} & 50.0 & 17.4 \\
    State-as-Text & 18.2 & 0.0 & 0.0 & 50.0 & 17.4 \\
    Action-FT & 9.1 & 0.0 & 0.0 & 50.0 & 13.0 \\
    \textbf{\ours} & \textbf{100.0} & \textbf{75.0} & \textbf{25.0} & \textbf{75.0} & \textbf{78.3} \\
    \bottomrule
  \end{tabular}
\end{table}

The explicit state of \ours can be edited during play. 
In \cref{tab:edits}, \ours carries out 78.3\% of the edits on unseen maps, against 13.0--21.7\% for the baselines.
Changes to the player's resources appear in the video almost always . Discontinuous edits are harder where a summoned monster or a teleport appears within 1\,s in only 25--75\% of cases, because a frozen backbone conditioned on clean history resists sudden changes of the scene. Such editability turns the model from a replay of what the player does into a game that can be programmed.

\subsection{Ablation study}
\label{sec:ablations}

\paragraph{Where the action enters.}
In \ours the action conditions only the state engine while previous works often inject action into video level~\cite{valevski2025gamengen,lin2026stateplay}.
\Cref{tab:action} tests the two alternatives where the action goes into the video backbone only or into both levels.
When only the video backbone receives the action, the engine must infer what the player does from the rendered video, and its rule states need to read the video too. Therefore in this alternative we change our selective correction into full correction.
The video still follows the action, but the state does not. Status agreement falls by about 9 points, and the engine's position error after 30\,s is about 2.1 times that of \ours, showing that inferring actions from rendered latents significantly limits the engine's state prediction. If both levels receive the action, the state does not improve and only the FVD is lower.
These results lead to our design where only state engine interacts with the action. This also aligns with our intuition in game development and programming.

\begin{table}[t]
  \centering
  \footnotesize
  \setlength{\tabcolsep}{2.6pt}
  \caption{\textbf{Ablation on where the action enters, training / unseen maps.}
    $\Delta$Pos.: paired change of the engine's position error after 30\,s against \ours.}
  \label{tab:action}
  \resizebox{\linewidth}{!}{%
  \begin{tabular}{@{}lcccc@{}}
    \toprule
    Action conditions & Status$\uparrow$ & Viol.$\downarrow$ & $\Delta$Pos.$\downarrow$ & FVD$\downarrow$ \\
    \midrule
    Video backbone only & 70.5 / 65.1 & \textbf{3.42} / 6.33 & $+$868 / $+$793 & 411 / 607 \\
    Engine and video backbone & 77.3 / 70.8 & 4.92 / 4.17 & $+$40 / $+$197 & \textbf{361} / \textbf{408} \\
    \textbf{Engine only (\ours)} & \textbf{79.3} / \textbf{74.3} & 5.92 / \textbf{1.92} & -- & 407 / 511 \\
    \bottomrule
  \end{tabular}}
\end{table}

\paragraph{The alignment loop.}
One of our main claims is that video level contains information that is not encoded in the state level hence the bi-directional alignment can help. We ablate the loop to verify this hypothesis. \Cref{tab:loop} removes the bottom-up link, so that the state still steers the video but the engine no longer sees what is rendered.
The engine then only rolls out along its own trajectory. Its ego-motion error rises by 2.36 on unseen maps and 1.65 on training maps, and after 30\,s its position is 58\% and 31\% further from the real player's. Status agreement falls by 4.1 and 0.7 points, and the enemy error rises by 0.09 on both. These results show that our bi-directional alignment indeed mitigates the neural engine's rollout error. 

\begin{table}[t]
  \centering
  \footnotesize
  \setlength{\tabcolsep}{4.5pt}
  \caption{\textbf{Ablation of the alignment loop.}
    Pos.: distance of the engine's position from the real player's after 30\,s (map units).}
  \label{tab:loop}
  \begin{tabular}{@{}llcccc@{}}
    \toprule
    Maps & Variant & Status$\uparrow$ & Enemy$\downarrow$ & Ego$\downarrow$ & Pos.$\downarrow$ \\
    \midrule
    \multirow{2}{*}{Training} & \textbf{\ours} & 79.3 & 0.361 & 7.63 & 745 \\
    & No bottom-up link & 78.6 & 0.446 & 9.28 & 979 \\
    \midrule
    \multirow{2}{*}{Unseen} & \textbf{\ours} & 74.3 & 0.557 & 6.85 & 701 \\
    & No bottom-up link & 70.2 & 0.645 & 9.21 & 1105 \\
    \bottomrule
  \end{tabular}
\end{table}

\begin{table}[t]
  \centering
  \footnotesize
  \setlength{\tabcolsep}{3.4pt}
  \caption{\textbf{Steering interfaces} (training / unseen maps).
    IoU: detected monsters against the state's boxes.}
  \label{tab:interfaces}
  \begin{tabular}{@{}lcccc@{}}
    \toprule
    Interfaces & Status$\uparrow$ & IoU$\uparrow$ & PSNR$\uparrow$ & FVD$\downarrow$ \\
    \midrule
    No CNN adapter & 80.7 / 65.3 & 0.05 / 0.05 & 18.67 / \textbf{17.90} & 395 / 405 \\
    No cross-attention & 97.9 / 91.1 & 0.21 / 0.30 & 18.25 / 17.61 & 347 / 399 \\
    No time slicing & 95.8 / 89.0 & 0.23 / 0.29 & \textbf{18.76} / 17.71 & 387 / 423 \\
    No LoRA & 98.7 / 91.3 & 0.22 / 0.30 & 18.20 / 17.52 & 326 / 367 \\
    \textbf{All three (\ours)} & \textbf{99.1} / \textbf{93.8} & \textbf{0.24} / \textbf{0.33} & 18.70 / 17.72 & \textbf{313} / \textbf{363} \\
    \bottomrule
  \end{tabular}
\end{table}

\paragraph{Steering interfaces.}
\Cref{tab:interfaces} removes one interface of \cref{sec:align} at a time. To test the influence of steering methods solely, we let the backbone render the true state for 8\,s, and the engine is not used here. All results are evaluated on checkpoints trained for 10k steps.
The CNN adapter matters most. Without its conditional maps, which also carry the digits of the status bar, status accuracy falls by 18.4 / 28.6 points, and monsters are no longer drawn in place.
The cross-attention adds 1.2 / 2.8 points to the maps, and removing its time slicing costs 3.3 / 4.8 points, as the counters change frame by frame.
Without LoRA, the number of trainable parameters is reduced from 128M to 81M. PSNR drops the most without LoRA, indicating that LoRA matters for generated video quality.

For further ablations on the engine's capacity, rollout loss and trajectory map, the state guidance, the decoder and the selective alignment please refer to the supplementary material for more details.

\subsection{Other backbones and games}
\label{sec:transfer}

\begin{table}[t]
  \centering
  \footnotesize
  \setlength{\tabcolsep}{2.6pt}
  \caption{\textbf{A second backbone and a second game.}
    Top: \doomstate closed loop on Wan2.2-TI2V-5B, training / Unseen maps.
    Bottom: WildWorld on both backbones at equal budget.}
  \label{tab:transfer}
  \resizebox{\linewidth}{!}{%
  \begin{tabular}{@{}lcccc@{}}
    \toprule
    \doomstate, 5B backbone & Status$\uparrow$ & Ego$\downarrow$ & Viol.$\downarrow$ & FVD$\downarrow$ \\
    \midrule
    Action-LoRA & 62.1 / 58.1 & 7.67 / 6.96 & 26.7 / 32.3 & 551 / 564 \\
    AVID & 29.5 / 27.3 & 9.31 / 9.25 & 71.3 / 83.9 & 799 / 968 \\
    State-as-Text & 65.7 / 59.8 & 10.41 / 9.63 & 26.3 / 27.6 & 700 / 650 \\
    Rule-Engine & 66.8 / 61.9 & 13.30 / 11.64 & \textbf{12.8} / 14.3 & 545 / 772 \\
    StatePlay & 59.7 / 53.0 & 8.73 / 8.66 & 14.4 / 17.8 & \textbf{235} / \textbf{519} \\
    \textbf{\ours} & \textbf{75.2} / \textbf{65.9} & \textbf{6.81} / \textbf{6.47} & 13.3 / \textbf{10.3} & 462 / 573 \\
    \midrule
    WildWorld & PSNR$\uparrow$ & SSIM$\uparrow$ & LPIPS$\downarrow$ & FVD$\downarrow$ \\
    \midrule
    Action-LoRA, 1.3B & 17.03 & 0.432 & \textbf{0.732} & 603 \\
    \textbf{\ours}, 1.3B & \textbf{17.44} & \textbf{0.446} & 0.756 & \textbf{533} \\
    Action-LoRA, 5B & 17.58 & 0.439 & \textbf{0.605} & 760 \\
    \textbf{\ours}, 5B & \textbf{18.01} & \textbf{0.461} & 0.652 & \textbf{569} \\
    \bottomrule
  \end{tabular}}
\end{table}

\paragraph{A 5B backbone.}
We apply the same recipe to Wan2.2-TI2V-5B~\cite{wan2025, wan22}, a backbone that was not trained with diffusion forcing. The same interfaces, widened to the backbone, give 163M trainable parameters.
\ours is again best on every state metric except violations on training maps, where Rule-Engine is lowest.
The metrics drop slightly compared with the 1.3B backbone, because the coarser autoencoder uses one token to represent $32{\times}32$ pixels and loses some small visual details. For example, the 4-pixel ammunition counters fall to 64--73\% accuracy, against 87--98\%.

\begin{figure}[t]
  \centering
  \includegraphics[width=\linewidth]{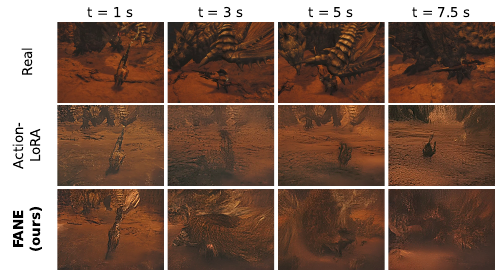}
  \caption{\textbf{WildWorld.} Rendering a photorealistic third-person game at equal budget with the 5B backbone.
    Rows: real video, Action-LoRA, \ours.}
  \label{fig:wildworld}
\end{figure}

\paragraph{A photorealistic third-person game.}
We also test our methods on another photorealistic third-person game and only video quality is evaluated. On WildWorld (\cref{tab:transfer}; \cref{fig:wildworld}), \ours improves over Action-LoRA at equal budget in PSNR, SSIM and FVD on both backbones , while Action-LoRA has the lower LPIPS. In \cref{fig:wildworld} the monster disappears in Action-LoRA's rollout while \ours is still aware of it. 

\subsection{Qualitative study}
\label{sec:qual}

\Cref{fig:doom} compares 30-second rollouts from the same first frame and the same recorded actions.
AVID loses the scene in the early stage, and its HUD stays unchanged. StatePlay stays clear for the whole rollout, in line with its low FVD, but its HUD is not tied to what happens. The ammunition rises from 41 to 48, and the player keeps playing at 0\% health, the kind of error counted as violations in \cref{tab:main}.
Rule-Engine has the ammunition of the real game at every time shown, since it executes the weapon rules on the recorded actions, but its player loses almost all health in a fight that the real player goes through unharmed.
\ours keeps the scene clear, and its counters only decrease.
The remaining difference from the real game is a fight that develops differently. Considering both visual and rule consistency, \ours behaves closest to the real game.

\section{Limitations and Discussion}
\label{sec:discussion}

\paragraph{Limitations.}
\ours is not real-time. Closed-loop generation runs at 0.35 frames per second on one GPU, and we leave few-step distillation for playable simulation~\cite{yin2025causvid,huang2025selfforcing} to future work. Without a map file, the backbone has to invent the parts of a map it has not seen, so the layout drifts from the real game; a state engine aware of the environment may help. Finally, the method needs logged states of the target game, and its state schema is still chosen by hand, which limits transfer to other games.

\paragraph{Discussion and future work.}
The formulation does not tie the engine to one game. A general state engine, such as a language model or a model pretrained on the state logs of many games, could be aligned with a video backbone in the same way, towards a single simulator of many games. A further level of memory could be added to improve consistency over longer horizons.

\section{Conclusion}
\label{sec:conclusion}

In this paper, we presented \ours, a hierarchical architecture that learns game simulation from both game state and video. The state engine steers the video with the quantities it holds exactly, and the video generator corrects the geometry that the state predicts.
We also release \doomstate, a benchmark with the exact state of every frame and exact counterfactual ground truth for state edits; on its unseen maps, \ours breaks the rules less often than a hand-written engine and video-only models, keeps the lowest errors in the number of visible monsters and in the camera motion, and carries out most state edits during play, while training only a lightweight state engine and adaptation modules to the video backbone.
These results show that a state level and a video level can complement each other, which we hope will inform the design of future game world models.

\section*{LLM Usage Statement}
We used Claude (Anthropic), through its coding and chat interfaces, in two parts of this work.
(i) Research code: it wrote and debugged most of the code for data collection, training, evaluation and analysis under the authors' specification and review. All reported numbers are produced by this code from the evaluation logs.
(ii) Writing: it polished the paper writing and edited parts of the text, tables and figure layouts from the authors' specifications. The authors revised the text and take responsibility for the full paper.

{
    \small
    \bibliographystyle{ieeenat_fullname}
    \bibliography{main}
}

\clearpage
\setcounter{page}{1}
\maketitlesupplementary
\appendix
\backrefsetup{disable}
\raggedbottom
\setcounter{table}{0}
\setcounter{figure}{0}
\setcounter{equation}{0}
\renewcommand{\thetable}{S\arabic{table}}
\renewcommand{\thefigure}{S\arabic{figure}}
\renewcommand{\theequation}{S\arabic{equation}}
\makeatletter
\renewcommand{\theHtable}{S\arabic{table}}
\renewcommand{\theHfigure}{S\arabic{figure}}
\renewcommand{\theHequation}{S\arabic{equation}}
\makeatother
\newcommand{\pair}[2]{#1\,/\,#2}

\section{Implementation details}
\label{sec:s_method}

\subsection{State and actions}
\label{sec:s_state}

The state $s_t$ of a frame has three fields.
The \emph{player} field includes position, normalized by the map's bounding box, velocity, the sine and cosine of the yaw, pitch, health, armor, the four ammunition counts, a one-hot weapon and an alive flag.
The \emph{entities} occupy 16 slots and each includes position relative to the player, distance, bearing, relative velocity, relative yaw, type, and flags for alive and visible.
Slots are filled with the visible entities first, then the living threats, then the rest, in order of distance from the player, nearest first.
The \emph{events} indicate what occurs during the current frame including shot, hit dealt, damage taken, pickup, kill, death and use.
The geometry state $\sgeo_t$ consists of the continuous values like positions, velocities and orientations of the player and the entities; the rule state $\srule_t$ consists of the discrete ones.
The action $a_t$ has 9 values per frame. Eight are binary buttons: move forward, move backward, strafe left, strafe right, attack, use, select the next weapon, and run. The ninth is a continuous turn from the mouse, the change of the yaw in degrees during the frame, which the game applies exactly. The action $a_t$ is executed between frames $t$ and $t+1$, so the engine reads $a_t$ when it predicts $s_{t+1}$.

\subsection{Neural state engine}
\label{sec:s_engine}

\paragraph{Architecture.}

A causal transformer $f_\phi$ serves as the neural state engine, which has 8 layers of width 512 with rotary position embeddings and a context of 64 frames.
The tokens of a frame are the player's state, the action, the events, one token per entity slot and a level token recording the player's recent trajectory. An extra 2-layer encoder aggregates heterogeneous tokens into a unified embedding $e_t$ for the neural engine.

To recover the predicted state $s_{t+1}$ from the embedding $h_t$, we add prediction heads after the transformer, one for each type of variable.
For the continuous variables, a Gaussian-mixture head outputs the motion of the player and of every entity, and Gaussian heads output the remaining values, such as velocity, pitch and relative yaw.
For the discrete variables, classification heads output the changes of health, armor and ammunition, categorical heads output the weapon and the entity type, and binary heads output the flags (present, alive, visible) and the events.

\paragraph{Trajectory map.}
A trajectory map is stored with the state engine to record the positions that have already been visited.
The level token encodes a $64{\times}64$ crop around the player and rotated to the heading, of an occupancy map that the engine draws from its own trajectory: cells that the player has walked over are free, and cells that the player pushed against without moving are blocked.
Frames that leave the 64-frame context are added to this map.

\paragraph{Vision corrector.}
The vision corrector consists of three convolutional layers and two layers of cross-attention with feed-forward blocks.
At every update, the convolutional layers compress each of the 4 most recently rendered latents $X_t$ into a $6{\times}8$ grid of tokens, to which positional embeddings are added.
The state tokens of frame $t$, which form the embedding $e_t$, attend to these tokens through the cross-attention and feed-forward layers, and the output is an incremental correction of the state that is initialized to zero.
An additional learned null token enables frames without video latents to take the same path.

\paragraph{Training.}
Training windows have 80 frames.
The negative log-likelihood is computed with teacher forcing over 64 frames.
Every 16 training steps a rollout loss (\cref{eq:roll}) is added with weight $\lambda=0.5$: after 8 teacher-forced frames the engine is unrolled on its own reparameterized samples for $H$ steps, with $H$ growing from 16 to 64.
The vision corrector is trained on the latents of the real video. The video is dropped for half of the frames, and the latents receive noise with $\sigma\sim U(0,0.3)$ to improve robustness. In a quarter of the windows the video only starts inside the window, as at the start of a rollout.
The engine and its vision corrector are both trained with AdamW, a learning rate of $3\times10^{-4}$, dropout 0.1, and 24k steps with a batch of 32 per GPU on 2 GPUs.

\subsection{Steering interfaces}
\label{sec:s_interfaces}

\paragraph{Backbone.}
SkyReels-V2-DF-1.3B~\cite{chen2025skyreels} has 30 blocks of width 1536.
The Wan~2.1 autoencoder~\cite{wan2025} has 16 latent channels and compresses 4 frames and $8{\times}8$ pixels into one latent position, so a frame of $384{\times}512$ pixels becomes a latent of $48{\times}64$, or 768 tokens after $2{\times}2$ patching.
Wan2.2-TI2V-5B~\cite{wan22} also has 30 blocks, of width 3072.
Its autoencoder has 48 latent channels and compresses 4 frames and $16{\times}16$ pixels into one latent position, so the same frame becomes a latent of $24{\times}32$, or 192 tokens after $2{\times}2$ patching: one token per $32{\times}32$ pixels.
Unlike the 1.3B backbone, it was trained with a clean first frame and not with diffusion forcing.
Both backbones are frozen and run in bfloat16 with the text prompt fixed.

\paragraph{State tokens and time-sliced cross-attention.}
A frame gives 18 state tokens of width 512: the player, the 16 entity slots and the events.
A 2-layer transformer fuses the 4 frames of a latent along the frame axis, and the result is mean-pooled and receives an embedding of the latent's index; $c_j$ in \cref{eq:xattn} denotes these 18 tokens.
New cross-attention layers sit in every second block, and their output is added to the output of the block's text cross-attention.
The time slicing is implemented without an attention mask: the latent axis is moved into the batch dimension, so that the 768 video tokens of latent $j$ attend to the 36 state tokens of latents $j-1$ and $j$, with a learned padding for $j=0$.
The output projection of every new layer is initialized to zero, so training starts from the pretrained model. 

\paragraph{Condition maps and CNN adapter.}
The condition maps draw the occupancy of the visible entities per category as projected cylinders with the nearest in front, the inverse depth of the nearest entity, the Pl\"ucker coordinates of the camera rays relative to the first frame of the window, and the digit shown in each cell of the HUD.
The adapter groups the 4 frames of a latent with a 3D convolution of kernel and stride 4 in time followed by two spatial convolutions.
Its output is added to the hidden states that enter blocks 0, 6, 12, 18 and 24.

\paragraph{LoRA.}
LoRA of rank 64 is applied to the query, key, value and output projections of the self-attention and the cross-attention of all 30 blocks.
\Cref{tab:s_params} lists the trainable parameters on both backbones.

\begin{table}[t]
  \centering
  \footnotesize
  \setlength{\tabcolsep}{5pt}
  \caption{\textbf{Trainable parameters} of \ours on the two backbones.}
  \label{tab:s_params}
  \begin{tabular}{@{}lrr@{}}
    \toprule
    Module & 1.3B & 5B \\
    \midrule
    LoRA, rank 64 & 47.2M & 94.4M \\
    State cross-attention, 15 blocks & 31.7M & 55.3M \\
    State encoder & 5.1M & 5.1M \\
    CNN adapter & 4.3M & 8.3M \\
    Neural state engine & 35.6M & 35.6M \\
    Vision corrector & 4.8M & 4.8M \\
    Total & 128.7M & 203.5M \\
    \bottomrule
  \end{tabular}
\end{table}

\paragraph{Training.}
The interfaces are trained with \cref{eq:fm} on windows of 13 latents.
The noise level is drawn uniformly and shifted with a shift of 5.
Half of the windows use an independent noise level per latent; the other half use a clean history prefix whose latents receive noise up to $\sigma=0.3$, which exposes the model to imperfect history.
The loss on the HUD rows has weight 1.
The conditioning state of a clip is a mixture of the logged state, the engine's one-step prediction or the logged state with Gaussian noise. The state is dropped entirely for 10\% of the clips, replaced by learned null tokens and empty condition maps.
We train for 30k steps with a batch of 2 per GPU on 4 GPUs and a learning rate of $10^{-4}$.
Both backbones are trained with this recipe.

\subsection{Other techniques for improving performance}
\label{sec:s_training}

\paragraph{Decoder.}
Following previous work~\cite{valevski2025gamengen}, we further finetune the decoder to make it better adapt to the imperfect latents generated by the backbone.
It is trained with L1 and LPIPS losses, first on the latents of real frames and then mostly on latents generated by our model, paired with the real frames.
Where a generated window is not pixel-aligned with the real frames, the L1 loss compares low-pass versions of the two.
For the adversarial loss, a per-frame patch discriminator~\cite{isola2017pix2pix} is trained with the hinge loss alongside the decoder, and its loss on the decoded frames is added to the decoder's loss with the adaptive weight of VQGAN~\cite{esser2021taming}.

\paragraph{Sampling.}
The 3 new latents of a chunk are denoised together in 20 Euler steps with a shift of 5, given 10 history latents at a noise level of 0.02.
The state guidance is $v=v_\varnothing+1.5\,(v_s-v_\varnothing)$, where $v_s$ and $v_\varnothing$ are the predictions with and without the state.
A rollout starts from one real frame; the engine receives the logged states of up to 64 frames before it.

\paragraph{Self-forcing fine-tuning.}
We find that self-forcing fine-tuning can further improve the video quality though it is not used in the main experiments~\cite{huang2025selfforcing}. After \ours's training stage, we continue to train our model for 10k steps in which 1 to 4 real latents are followed by 6 latents that the model generates itself, in 2 chunks with 4 Euler steps and without gradient, before the real targets. \cref{sec:s_sf} reports its effect.

\section{\doomstate and the evaluation protocol}
\label{sec:s_bench}

\subsection{Data}
\label{sec:s_data}

\doomstate is recorded with ViZDoom~\cite{kempka2016vizdoom} at 2 game tics per frame.
\Cref{tab:s_data} lists the splits.
The 15 training maps are 12 maps of Freedoom: Phase~2 (MAP01--03, 05--07, 09--11, 13, 14 and 17) and the ViZDoom scenarios deathmatch, health gathering supreme and defend the center; the 5 unseen maps are MAP04, 08, 12 and 19 and the scenario defend the line.
Three policies play the episodes: pretrained Sample Factory agents~\cite{petrenko2020samplefactory}, sticky random actions, and waypoint routes that revisit places, during which the player cannot die.
An episode lasts at most 120\,s.
Results on training maps use new episodes with new seeds, which are never trained on.

\begin{table}[t]
  \centering
  \footnotesize
  \setlength{\tabcolsep}{4pt}
  \caption{\textbf{\doomstate splits.} Policies: agent / random / route episodes.}
  \label{tab:s_data}
  \begin{tabular}{@{}lrrrl@{}}
    \toprule
    Split & Maps & Episodes & Hours & Policies \\
    \midrule
    Training & 15 & 2{,}963 & 60.9 & 1{,}255 / 1{,}061 / 647 \\
    Test, unseen maps & 5 & 505 & 6.2 & 217 / 202 / 86 \\
    Test, training maps & 15 & 300 & 6.6 & 41 / 40 / 219 \\
    \bottomrule
  \end{tabular}
\end{table}

\subsection{Tasks}
\label{sec:s_tasks}

\Cref{tab:s_tasks} lists all tasks.
The items of a task (episode, start frame and sampling seed) were fixed before any model was evaluated, every task exists on both training maps and on unseen maps.

\begin{table}[t]
  \centering
  \footnotesize
  \setlength{\tabcolsep}{3.5pt}
  \caption{\textbf{Tasks.} Items on training / unseen maps. In the tasks driven by the true state, the engine is not used.}
  \label{tab:s_tasks}
  \begin{tabular}{@{}lllr@{}}
    \toprule
    Task & Driven by & Items & Length \\
    \midrule
    Closed-loop rollout & recorded actions & 24 / 24 & 30\,s \\
    State editing & true state, edit at 2\,s & 24 / 24 & 6\,s \\
    Rendering from the true state & true state & 32 / 32 & 8\,s \\
    Revisit routes & recorded actions & 16 / 16 & 60\,s \\
    WildWorld & logged state & 24 & 8\,s \\
    \bottomrule
  \end{tabular}
\end{table}

The 24 edits of a split are 12 that give ammunition, health or armor, 4 that take ammunition, 4 that summon a monster and 4 that warp the player.
For an edit, the engine of the game is run twice from the same point, with and without the edit, which gives the two reference videos.

\subsection{Evaluator}
\label{sec:s_evaluator}

The evaluator is fitted on real frames of training episodes only and never sees generated video.
The \emph{HUD reader} matches templates per digit cell.
The \emph{detector}, a Faster R-CNN with a MobileNetV3 backbone trained on the boxes of the game's label buffer is used to detect the enemies.
The \emph{ego-motion regressor}, a CNN on pairs of consecutive frames, gives the predictions of map units for the forward, the sideways displacement and angles for the turn.
The \emph{event classifier}, a 3D CNN on windows of 4 frames, is used by the rule checks.

\subsection{WildWorld}
\label{sec:s_wildworld}

We use a 40-hour subset of WildWorld~\cite{li2026wildworld} with 226 samples, of which 202 (36.0\,h) are used for training and 24 (4.2\,h) for testing.
The split is by sample, so the test samples show regions that also occur in training.
The videos are converted to 17.5 frames per second and center-cropped to $384{\times}512$ pixels.
The state holds the tracked character and 4 entity slots, with events derived from changes of the health points; the action is the character's motion and weapon.
The camera pose and intrinsics are logged per frame and used for the condition maps, in which every monster is drawn with the same placeholder size.
There is no HUD and no evaluator for this game, so only image metrics are reported.
The dataset is used under its license for non-commercial research.

\section{Baselines}
\label{sec:s_baselines}

All baselines use the same data, training windows, 30k steps, sampler and fine-tuned decoder as \ours, with guidance of 1.5 on their own condition unless noted.

\paragraph{Action-LoRA and Action-FT.}
The action of every frame becomes one token, and the tokens enter through the same time-sliced cross-attention as our state tokens.
Action-LoRA adds LoRA of rank 32.
Action-FT fine-tunes the whole backbone instead, with a learning rate of $10^{-5}$.

\paragraph{State-as-Text.}
The state is written as text and enters through the backbone's own text cross-attention, with LoRA of rank 32.
In closed loop the text is written from the state of our final engine, which reads the video that this baseline renders.

\paragraph{AVID.}
We port the configuration of AVID~\cite{rigter2025avid}: the frozen backbone is run as a black box, and a 3D U-Net adapter reads the noisy latents and the backbone's prediction and mixes its own prediction in through a learned mask.
We adapt it to flow matching and to one noise level per latent.
As in its official implementation, it is trained without action dropout, sampled without guidance, and evaluated with an exponential moving average of its weights.

\paragraph{Rule-Engine.}
The learned engine is replaced by hand-written rules of the game: the player's physics, the weapons, hit-scan damage, the monsters' chase and attack behavior, armor, pickups and the events.
The interfaces and the decoder are those of \ours.
Like our engine, it starts from the state of the first frame and has no map file.

\paragraph{StatePlay.}
StatePlay~\cite{lin2026stateplay} generates the video and a state vector jointly, with a state path and a state--video cross-attention in every block, and is fully fine-tuned.
We port it to both backbones with a state of 13 values (health, armor, the selected and the four ammunition counts, the weapon, the position, the yaw and the number of visible monsters).
The official implementation generates 101 frames at once; for longer rollouts we generate chunks of 12 latents, each starting from the last generated frame, with 30 steps and a text guidance of 5 as in the official implementation.
In closed loop, the first chunk receives the true state of the first frame, and every later chunk the state that StatePlay itself predicted.

\section{Additional results on \doomstate}
\label{sec:s_results}

\subsection{Rendering from the true state}
\label{sec:s_t1}

In this task the true state of the game drives the model for 8\,s, so the engine plays no part and the task measures how well a model can be controlled.
The action-conditioned baselines receive the recorded actions.
StatePlay is not part of this task: it generates its state jointly with the video, so the true state can be given to it only for the first frame.
\Cref{tab:s_t1} shows that \ours draws the HUD of the given state almost exactly (99.6\% and 97.3\%) and places the monsters where the state has them, while every baseline stays below 80\% and below an IoU of 0.1; \ours also has the best LPIPS and FVD.

\begin{table}[t]
  \centering
  \footnotesize
  \setlength{\tabcolsep}{4pt}
  \caption{\textbf{Rendering from the true state}.
    Status: accuracy of the HUD against the given state (\%); Enemy and IoU: detected monsters against the given state.}
  \label{tab:s_t1}
  \begin{tabular}{@{}lcccccc@{}}
    \toprule
    Method & Status$\uparrow$ & Enemy$\downarrow$ & IoU$\uparrow$ & PSNR$\uparrow$ & LPIPS$\downarrow$ & FVD$\downarrow$ \\
    \midrule
    \multicolumn{7}{@{}l}{\emph{Training maps}} \\
    Action-LoRA & 54.3 & 0.527 & 0.07 & 16.65 & 0.626 & 387 \\
    AVID & 79.4 & 0.454 & 0.05 & 17.87 & 0.531 & 738 \\
    Action-FT & 76.5 & 0.488 & 0.05 & 17.98 & 0.552 & 311 \\
    State-as-Text & 76.2 & 0.426 & 0.06 & 17.65 & 0.615 & 437 \\
    \textbf{\ours} & \textbf{99.6} & \textbf{0.388} & \textbf{0.24} & \textbf{18.80} & \textbf{0.493} & \textbf{299} \\
    \midrule
    \multicolumn{7}{@{}l}{\emph{Unseen maps}} \\
    Action-LoRA & 53.5 & 0.621 & 0.04 & 16.58 & 0.633 & 460 \\
    AVID & 63.4 & 0.566 & 0.04 & 17.08 & 0.543 & 629 \\
    Action-FT & 57.7 & 0.605 & 0.06 & 16.81 & 0.600 & 421 \\
    State-as-Text & 60.8 & 0.603 & 0.07 & 16.83 & 0.659 & 520 \\
    \textbf{\ours} & \textbf{97.3} & \textbf{0.463} & \textbf{0.34} & \textbf{17.43} & \textbf{0.535} & \textbf{339} \\
    \bottomrule
  \end{tabular}
\end{table}

\subsection{Revisit routes}
\label{sec:s_t3}

The revisit routes are 60-second closed-loop rollouts on episodes whose route returns to places visited before (\cref{tab:s_t3}).
A revisit pair is two frames of the same rollout, at least 3\,s apart, whose poses agree within 32 map units and $10^\circ$. A real episode has 64 pairs in our test set.
The revisit LPIPS is the LPIPS between the two frames of a pair, so it measures whether a place still looks the same when the player returns to it.
\ours keeps the HUD with the game better than Action-LoRA, Action-FT, StatePlay, State-as-Text and AVID, with 4.1 and 6.1 violations per minute against 12--82 for them.
Its revisit LPIPS is the lowest on training maps and ties with AVID on unseen maps.
Rule-Engine keeps the HUD about as well and has fewer violations, but only 27 and 20 revisit pairs per item are found in its rollouts, against 64 in the real episodes.

\begin{table}[t]
  \centering
  \footnotesize
  \setlength{\tabcolsep}{3pt}
  \caption{\textbf{Revisit routes}. Pairs: revisit pairs found per item.}
  \label{tab:s_t3}
  \begin{tabular}{@{}lcccc@{}}
    \toprule
    Method & Status$\uparrow$ & Viol.$\downarrow$ & Rev.\ LPIPS$\downarrow$ & Pairs \\
    \midrule
    \multicolumn{5}{@{}l}{\emph{Training maps}} \\
    Action-LoRA & 52.1 & 26.81 & 0.560 & 55.4 \\
    AVID & 67.0 & 46.94 & 0.479 & 61.7 \\
    Action-FT & 67.2 & 12.31 & 0.477 & 51.9 \\
    StatePlay & 70.0 & 46.25 & 0.470 & 57.7 \\
    State-as-Text & 55.6 & 29.94 & 0.421 & 55.1 \\
    Rule-Engine & 72.6 & \textbf{1.88} & 0.518 & 27.4 \\
    \textbf{\ours} & \textbf{75.9} & 4.06 & \textbf{0.409} & 61.1 \\
    \midrule
    \multicolumn{5}{@{}l}{\emph{Unseen maps}} \\
    Action-LoRA & 32.7 & 23.69 & 0.589 & 50.2 \\
    AVID & 43.1 & 53.94 & \textbf{0.495} & 55.1 \\
    Action-FT & 40.8 & 22.38 & 0.560 & 47.2 \\
    StatePlay & 43.1 & 82.38 & 0.536 & 34.2 \\
    State-as-Text & 32.6 & 41.44 & 0.510 & 55.0 \\
    Rule-Engine & \textbf{53.7} & \textbf{4.19} & 0.522 & 20.1 \\
    \textbf{\ours} & 52.8 & 6.12 & 0.497 & 59.5 \\
    \bottomrule
  \end{tabular}
\end{table}

\subsection{Self-forcing}
\label{sec:s_sf}

\Cref{tab:s_sf} compares \ours with its self-forcing variant (\cref{sec:s_training}).
The self-forcing stage mainly improves the video quality of long rollouts: in the 30-second closed loop the FVD falls from 407 to 263 on training maps and from 511 to 406 on unseen maps, and the LPIPS falls on both.
The state metrics change little and not always for the better: the status agreement is 79.6\% and 72.5\%, against 79.3\% and 74.3\% without the stage.
When rendering from the true state for 8\,s, the status accuracy is unchanged, the IoU of the monsters is slightly higher, and the FVD improves on training maps only.

\begin{table}[t]
  \centering
  \footnotesize
  \setlength{\tabcolsep}{3.2pt}
  \caption{\textbf{Self-forcing fine-tuning} (training / unseen maps).
    \ours as in the main paper and after the self-forcing stage, when rendering from the true state and in the closed-loop rollout.}
  \label{tab:s_sf}
  \begin{tabular}{@{}lcccc@{}}
    \toprule
    \multicolumn{5}{@{}l}{\emph{Rendering from the true state, 8\,s}} \\
    Method & Status$\uparrow$ & IoU$\uparrow$ & LPIPS$\downarrow$ & FVD$\downarrow$ \\
    \midrule
    \ours & 99.6 / 97.3 & 0.24 / 0.34 & 0.493 / 0.535 & 299 / \textbf{339} \\
    \ours + self-forcing & \textbf{99.7} / \textbf{97.5} & \textbf{0.25} / \textbf{0.38} & \textbf{0.465} / \textbf{0.524} & \textbf{260} / 356 \\
    \midrule
    \multicolumn{5}{@{}l}{\emph{Closed-loop rollout, 30\,s}} \\
    Method & Status$\uparrow$ & Viol.$\downarrow$ & LPIPS$\downarrow$ & FVD$\downarrow$ \\
    \midrule
    \ours & 79.3 / \textbf{74.3} & 5.92 / \textbf{1.92} & 0.549 / 0.589 & 407 / 511 \\
    \ours + self-forcing & \textbf{79.6} / 72.5 & \textbf{3.50} / 2.00 & \textbf{0.518} / \textbf{0.567} & \textbf{263} / \textbf{406} \\
    \bottomrule
  \end{tabular}
\end{table}

\section{Additional ablations}
\label{sec:s_ablations}

\subsection{Alignment loop and engine}
\label{sec:s_abl_engine}

The ablations of this subsection focus on the engine and the loop.
All of them use the interfaces of the main paper. Every variant is compared with \ours over the same items (\cref{tab:s_engine}).

\begin{table}[t]
  \centering
  \footnotesize
  \setlength{\tabcolsep}{7pt}
  \caption{\textbf{Loop and engine ablations} in the closed-loop rollout.}
  \label{tab:s_engine}
  \begin{tabular}{@{}lcccc@{}}
    \toprule
    Variant & Status$\uparrow$ & Enemy$\downarrow$ & Ego$\downarrow$ & Viol.$\downarrow$ \\
    \midrule
    \multicolumn{5}{@{}l}{\emph{Training maps}} \\
    \textbf{\ours} & 79.3 & 0.361 & 7.63 & 5.92 \\
    No bottom-up link & 78.6 & 0.446 & 9.28 & 2.33 \\
    No rollout loss & 80.4 & 0.521 & 7.81 & 3.50 \\
    10M engine & 74.6 & 0.529 & 8.03 & 2.83 \\
    No trajectory map & 76.1 & 0.480 & 7.56 & 3.75 \\
    Full correction & 74.7 & 0.472 & 9.00 & 2.00 \\
    \midrule
    \multicolumn{5}{@{}l}{\emph{Unseen maps}} \\
    \textbf{\ours} & 74.3 & 0.557 & 6.85 & 1.92 \\
    No bottom-up link & 70.2 & 0.645 & 9.21 & 2.83 \\
    No rollout loss & 76.7 & 0.782 & 6.75 & 3.83 \\
    10M engine & 71.7 & 0.670 & 7.70 & 3.25 \\
    No trajectory map & 67.0 & 0.782 & 7.62 & 5.25 \\
    Full correction & 67.0 & 0.485 & 8.17 & 4.00 \\
    \bottomrule
  \end{tabular}
\end{table}

\begin{table}[t]
  \centering
  \footnotesize
  \setlength{\tabcolsep}{5pt}
  \caption{\textbf{State of the engine against the real game} in the closed-loop rollout, measured without the video.
    State: agreement of the seven HUD counters in the engine's state with those of the game (\%); Health: the health counter alone.
    Pos.: distance between the engine's position and that of the real player, averaged over the rollout (map units).
    Step: error of the displacement per frame; the real player moves 12.0 and 11.3 units per frame.
    Enemies: error in the number of visible monsters.}
  \label{tab:s_engine_err}
  \begin{tabular}{@{}lccccc@{}}
    \toprule
    Variant & State$\uparrow$ & Health$\uparrow$ & Pos.$\downarrow$ & Step$\downarrow$ & Enemies$\downarrow$ \\
    \midrule
    \multicolumn{6}{@{}l}{\emph{Training maps}} \\
    \textbf{\ours} & 76.1 & 45.4 & 299 & 7.23 & 0.529 \\
    No bottom-up link & 77.2 & 39.7 & 345 & 7.04 & 0.521 \\
    No rollout loss & 80.2 & 43.6 & 326 & 7.20 & 0.642 \\
    10M engine & 74.3 & 49.7 & 326 & 7.25 & 0.567 \\
    No trajectory map & 75.7 & 47.6 & 339 & 7.37 & 0.557 \\
    Full correction & 74.6 & 40.9 & 411 & 7.96 & 0.597 \\
    \midrule
    \multicolumn{6}{@{}l}{\emph{Unseen maps}} \\
    \textbf{\ours} & 71.4 & 30.4 & 393 & 8.45 & 0.771 \\
    No bottom-up link & 71.5 & 25.6 & 440 & 8.48 & 0.740 \\
    No rollout loss & 76.9 & 28.2 & 395 & 8.87 & 1.043 \\
    10M engine & 71.9 & 24.4 & 363 & 8.46 & 0.857 \\
    No trajectory map & 67.3 & 26.3 & 414 & 8.94 & 0.981 \\
    Full correction & 67.0 & 24.7 & 434 & 9.28 & 0.675 \\
    \bottomrule
  \end{tabular}
\end{table}

\paragraph{Capacity, rollout loss and trajectory map.}
Removing the trajectory map costs the most status agreement on unseen maps, from 74.3\% to 67.0\%, and raises the enemy error there from 0.557 to 0.782 and the ego-motion error from 6.85 to 7.62.
On training maps its effect is smaller: the status agreement falls from 79.3\% to 76.1\% and the enemy error rises from 0.361 to 0.480.
Without the rollout loss the enemy error rises on both splits (0.521 against 0.361, and 0.782 against 0.557), while the status agreement is not reduced.
An engine of about 10M parameters (width 256) loses 4.7 and 2.6 points of status agreement and has a higher enemy error and ego-motion error on both splits (0.529 and 8.03 against 0.361 and 7.63 on training maps, 0.670 and 7.70 against 0.557 and 6.85 on unseen maps).

\paragraph{Engine error.}
The metrics above are measured on the video.
\Cref{tab:s_engine_err} compares the state that the engine itself holds during the closed loop with the state of the real game.
Some deviation is expected, because the behaviour of the monsters and the hits are random and differ from the recording: the health counter alone agrees in 24--50\% of the frames for every variant.
The HUD that the renderer draws agrees with the engine's state to 96.7\% and 98.4\%, so the gap between the HUD of \ours and the real game is the deviation of its engine instead of a limit of the rendering.
Over the rollout, the engine of \ours is on average 299 and 393 map units from the real player; without the bottom-up link the distance is 345 and 440 units, and without the trajectory map 339 and 414.
Without the rollout loss the engine's error in the number of visible monsters rises on both splits (0.642 and 1.043 against 0.529 and 0.771), although its HUD counters agree with the game more often.
The movement error per frame, 7.2 and 8.5 units for \ours, is large against the 12.0 and 11.3 units that the real player moves per frame; it arises mainly where the player is stopped abruptly by a wall and from the estimate of the velocity.

\subsection{Selective alignment}
\label{sec:s_selective}

\ours lets the video correct only the geometry state and predicts the rule state without the video.
The last row of \cref{tab:s_engine,tab:s_engine_err} removes this restriction on the same test items: the vision corrector may change every field of the state (full correction).
The video then pulls the engine away from the game.
The mean position error of the engine rises from 299 to 411 units on training maps and from 393 to 434 on unseen maps, and the ego-motion error of the video from 7.63 to 9.00 and from 6.85 to 8.17.
The status agreement is 4.6 and 7.3 points lower, and the enemy error is higher on training maps and lower on unseen maps.

\begin{table}[t]
  \centering
  \footnotesize
  \setlength{\tabcolsep}{2.8pt}
  \caption{\textbf{State guidance and decoder} when rendering from the true state. \ours: guidance scale 1.5 and the fine-tuned decoder, as in \cref{tab:s_t1}.}
  \label{tab:s_render}
  \begin{tabular}{@{}lcccccc@{}}
    \toprule
    Variant & Status$\uparrow$ & Enemy$\downarrow$ & IoU$\uparrow$ & PSNR$\uparrow$ & LPIPS$\downarrow$ & FVD$\downarrow$ \\
    \midrule
    \multicolumn{7}{@{}l}{\emph{Training maps}} \\
    \textbf{\ours} & 99.6 & 0.388 & 0.24 & 18.80 & 0.493 & 299 \\
    Guidance scale 1.0 & 99.5 & 0.409 & 0.22 & 19.16 & 0.508 & 332 \\
    Guidance scale 2.0 & 99.7 & 0.377 & 0.27 & 18.27 & 0.493 & 307 \\
    Guidance scale 3.0 & 99.7 & 0.362 & 0.31 & 17.48 & 0.498 & 329 \\
    Frozen decoder & 99.6 & 0.414 & 0.26 & 18.88 & 0.539 & 401 \\
    \midrule
    \multicolumn{7}{@{}l}{\emph{Unseen maps}} \\
    \textbf{\ours} & 97.3 & 0.463 & 0.34 & 17.43 & 0.535 & 339 \\
    Guidance scale 1.0 & 95.2 & 0.488 & 0.28 & 17.98 & 0.549 & 422 \\
    Guidance scale 2.0 & 97.6 & 0.419 & 0.40 & 16.83 & 0.537 & 328 \\
    Guidance scale 3.0 & 97.7 & 0.423 & 0.43 & 15.57 & 0.548 & 379 \\
    Frozen decoder & 97.3 & 0.483 & 0.31 & 17.46 & 0.575 & 397 \\
    \bottomrule
  \end{tabular}
\end{table}

\begin{figure*}[t]
  \centering
  \includegraphics[width=\linewidth]{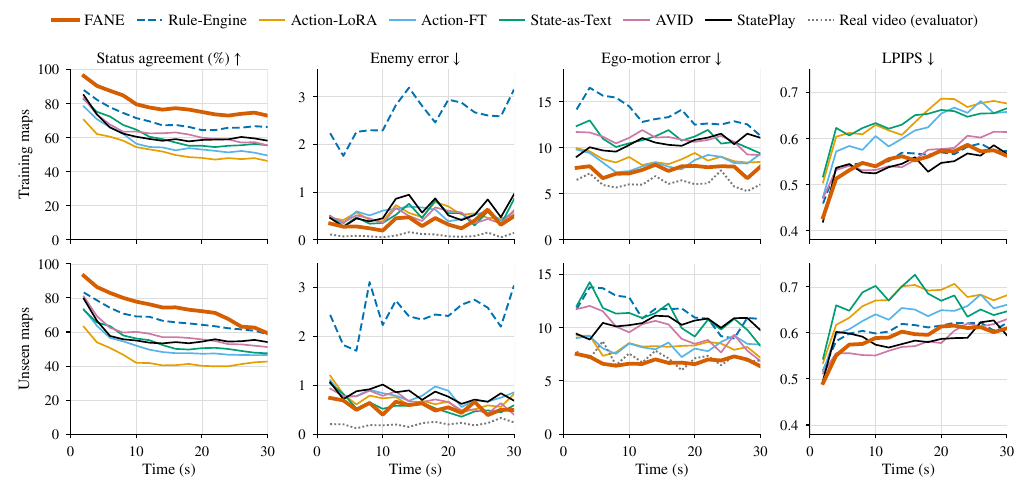}
  \caption{\textbf{Closed-loop metrics over the rollout.}
    The metrics of \cref{tab:main} during the 30\,s of the closed loop (top: training maps, bottom: unseen maps).
    Real video: the evaluator applied to the real video, the error floor of the metric.}
  \label{fig:s_curves}
\end{figure*}

\subsection{State guidance and decoder}
\label{sec:s_abl_render}

\Cref{tab:s_render} varies the sampler and the decoder when rendering from the true state; the reference is \ours as in \cref{tab:s_t1}.
Without guidance (scale 1.0) the model loses 2.1 points of status accuracy on unseen maps, places the monsters worse (IoU 0.28 against 0.34) and has a higher FVD (332 and 422 against 299 and 339), while its PSNR is higher (19.16 and 17.98 against 18.80 and 17.43).
Scales of 2.0 and 3.0 raise the IoU further (0.40 and 0.43 on unseen maps) but cost 0.6 and 1.9\,dB of PSNR there (0.5 and 1.3\,dB on training maps), and at 3.0 the FVD is higher on both splits; 1.5 is the balance.
The original, frozen decoder gives the same status accuracy, a slightly lower IoU on unseen maps (0.31 against 0.34) and worse images: its LPIPS is 0.539 and 0.575 against 0.493 and 0.535, and its FVD 401 and 397 against 299 and 339.

\section{Qualitative analysis}
\label{sec:s_qual}

\subsection{Metrics over the rollout}
\label{sec:s_curves}

\Cref{fig:s_curves} resolves the closed-loop results of \cref{tab:main} over the time of the rollout, from the same runs as the table.

\paragraph{Status.}
The status agreement of \ours is the highest from the first window on: 96.2\% on training maps and 93.1\% on unseen maps, against 70.5--87.9\% and 63.1--83.2\% for the baselines.
Every method loses most of its agreement during the first 10\,s and changes little afterwards, so the margin of \ours opens in the first seconds instead of building up over the rollout.
On training maps \ours stays 11--17 points above the best of Action-LoRA, Action-FT, State-as-Text, AVID and StatePlay, and 7--11 points above Rule-Engine, for the whole 30\,s.
On unseen maps the margins are 12--20 and 7--10 points up to 22\,s.

\paragraph{Ego-motion and enemies.}
The ego-motion error of \ours stays flat, at 6.7--8.1 map units on training maps and 6.4--7.5 on unseen maps, near the error floor of the evaluator.
It is the lowest of all methods in 29 of the 30 windows.
Rule-Engine has 11.3--16.5 and 9.0--13.8 units, and StatePlay 9.0--11.5 and 8.9--11.1.
About enemy error, \ours has 0.19--0.64 on training maps and 0.39--0.74 on unseen maps, the lowest value in 12 and 8 of the 15 windows, against a floor of 0.05--0.16 and 0.12--0.33.

\paragraph{LPIPS.}
The LPIPS against the real frames rises for every method in the first seconds and slowly afterwards.
From 10\,s on, \ours, Rule-Engine, AVID and StatePlay average 0.55--0.58 on training maps and 0.59--0.61 on unseen maps, and Action-LoRA, Action-FT and State-as-Text 0.64--0.66 and 0.65--0.69.
The image metric therefore does not separate \ours from AVID and StatePlay at any time of the rollout, while the state metrics do.

\begin{figure*}[t]
  \centering
  \includegraphics[width=\linewidth]{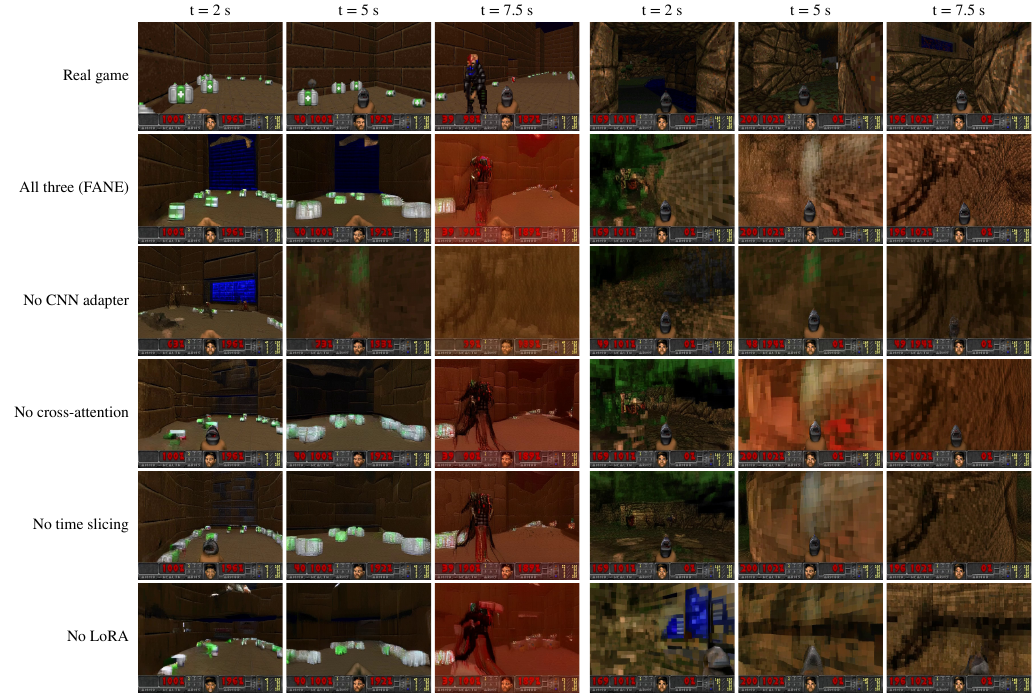}
  \caption{\textbf{Frames of the interface ablations} of \cref{tab:interfaces}.
    Every model renders the true state for 8\,s from the same first frame.
    Left: an item with a monster in view. Right: an item in which the ammunition changes by a pickup and by shots.}
  \label{fig:s_interfaces}
\end{figure*}

\subsection{Effect of the steering interfaces}
\label{sec:s_fig_interfaces}

\Cref{fig:s_interfaces} shows frames of the interface ablations of \cref{tab:interfaces}.
As in that table, every model renders the true state for 8\,s. The two items were selected among the 16 items with saved videos: one with a monster in view, and one in which the ammunition rises from 169 to 200 by a pickup and falls to 196 by shots.

\paragraph{CNN adapter.}
Without the CNN adapter the HUD is wrong in all six frames: the health reads 63\%, 73\% and 39\% where the game shows 100\%, 100\% and 98\%, and the ammunition 49, 48 and 49 instead of 169, 200 and 196.
The scene is lost as well.
After 2\,s the first item shows a blurred wall without the supplies and without the monster.
This agrees with \cref{tab:interfaces}, where this variant loses the most status accuracy and has an IoU of 0.05.

\paragraph{Cross-attention, time slicing and LoRA.}
The other four models draw the ammunition, the health and the armor of the second item correctly at all three times, including the pickup and the shots, and those of the first item at 2\,s and 5\,s.
All four draw a monster on the left at 7.5\,s, where the game has one, though as a smeared figure.
They differ in the image.
Without LoRA the frames have the least detail: the walls of the second item are coarse blocks and the weapon loses its shape, in line with the lowest PSNR of the table.
Without the cross-attention the second item shows colored patches at 5\,s that no other model has.
The frames without time slicing do not differ visibly from the reference.

\end{document}